\documentclass[10pt,conference,letterpaper]{IEEEtran}
\usepackage{times,amsmath,epsfig}
\usepackage{graphicx}
\usepackage{subfigure}
\usepackage{epsfig}
\usepackage{algorithm,algorithmic}
\usepackage{multirow}
\usepackage{color}
\usepackage{amssymb,amsfonts}

\renewcommand{\vec}[1]{\mathbf{#1}}
\usepackage{cite,citesort}

\title{Precision Enhancement of 3D Surfaces from Multiple Compressed Depth Maps}
\author{%
{Pengfei Wan{\small $^{\#}$}, Gene Cheung{\small $^{*}$},
Philip A. Chou{\small $^{\$}$}, Dinei Florencio{\small $^{\$}$},
Cha Zhang{\small $^{\$}$}, Oscar C. Au{\small $^{\#}$} }%
\vspace{1.6mm}\\
\fontsize{10}{10}\selectfont\itshape
$^{\#}$\,Hong Kong University of Science and Technology\\
Clear Water Bay, Kowloon, Hong Kong\\
\fontsize{9}{9}\selectfont\ttfamily\upshape
\{leoman, eeau\}@ust.hk%
\vspace{1.2mm}\\
\fontsize{10}{10}\selectfont\rmfamily\itshape
$^{*}$\,National Institute of Informatics\\
2-1-2, Hitotsubashi, Chiyoda-ku, Tokyo, 101-8430, Japan\\
\fontsize{9}{9}\selectfont\ttfamily\upshape
cheung@nii.ac.jp
\vspace{1.2mm}\\
\fontsize{10}{10}\selectfont\rmfamily\itshape
$^{\$}$\,Microsoft Research\\
One Microsoft Way, Redmond, WA 98052-6399, USA\\
\fontsize{9}{9}\selectfont\ttfamily\upshape
\{pachou, dinei, chazhang\}@microsoft.com
}
\begin{document}
\maketitle



\begin{abstract}
In texture-plus-depth representation of a 3D scene, depth maps from 
different camera viewpoints are typically lossily 
compressed via the classical transform coding / coefficient quantization 
paradigm. In this paper we propose to reduce distortion of 
the decoded depth maps due to quantization. The key observation
is that depth maps from different viewpoints constitute multiple 
descriptions (MD) of the same 3D scene. Considering the MD jointly, 
we perform a POCS-like iterative procedure to project a reconstructed signal
from one depth map to the other and back, so that the converged 
depth maps have higher precision than the original quantized versions.  
\end{abstract}

%


\section{Introduction}
\label{sec:intro}
Texture-plus-depth is now a popular format for dynamic 3D 
scene representation, where texture and depth maps from multiple viewpoints
are captured and compressed. Multiview depth maps are often corrupted by acquisition noise; in \cite{wenxiu13}, we denoised and compressed depth maps simultaneously in a rate-distortion optimal manner. Depth maps are also constrained by sensor's limited acquisition bit-depth; in \cite{ivmsp13}, we enhanced the precision of depth samples by considering multiple depth maps jointly. In contrast, in this paper we focus on reducing the distortion of depth maps caused by lossy transform coding.




The key observation in our work is that depth maps from different 
viewpoints can be interpreted as \textit{multiple descriptions} (MD) of 
the same 3D scene. 
Thanks to this representation redundancy, one can enhance the quality
of the reconstructed depth maps by considering multiple descriptions
jointly. In particular, we propose an iterative procedure
inspired by \textit{projections onto convex sets} 
(POCS)~\cite{phil99_pocs}, 
where a reconstructed 2D signal from the left compressed depth map is first
back-projected to the 3D space, then projected to the right depth map.
Then the projected signal is appropriately clipped to satisfy the 
quantization bin constraints of the right depth map. This transformation
from left to right depth map constitutes one projection, and the second
projection from right to left depth map is performed similarly,
and the two projections are repeated until convergence.
We show experimentally that the converged signal has higher precision 
than the original quantized version of individual depth map.


\section{System Overview}
\label{sec:system}

For simplicity, we consider a scenario where a static 3D scene is captured 
by depth maps of only two views (left and right). Each depth map is 
compressed using JPEG; $8 \times 8$ DCT transform coefficients 
are quantized and transmitted. At the decoder, given quantized 
transform coefficients, our goal is to reconstruct depth maps with 
enhanced precision.

Let $\vec{x}$ be a vectorized representation of an original 
$8 \times 8$ pixel block in the left depth map. The corresponding 
transform coefficient vector is $\vec{y} = \mathbf{T} \vec{x}$, where
$\mathbf{T}$ is the linear DCT transform operator. Scalar quantization
of each coefficient $k$, $y_k$, means mapping of $y_k$ to quantization
bin $\mathcal{B}_k$ of centroid $q(y_k)$ and width $\Delta$, where
\begin{equation}
y_k \in \left[ q(y_k) - \Delta/2, q(y_k) + \Delta/2 \right] 
\stackrel{\mathrm{def}}{=} \mathcal{B}_k
\end{equation}
 
Only quantization bin indices are encoded, so at decoder, only
quantization bins $\mathcal{B}_k$'s are known. 
We can hence write the quantization bin constraints for the 
reconstructed coefficients $\mathbf{y}^l$ for a $8 \times 8$ pixel block 
in the left depth map as:
\begin{equation}
\mathbf{y}^l \in \mathcal{S}^l \stackrel{\mathrm{def}}{=} 
\left\{ 
\mathbf{y} ~|~ y_k \in \mathcal{B}_k \right\}
\label{eq:quan}
\end{equation}

Geometrically, $\mathcal{S}^l$ is a hyper-cube representing the range
of transform coefficients for a quantized block in the left depth map. 
A sequence of $\mathcal{S}^l(\vec{b})$'s for different blocks $\vec{b}$ in the left
map thus defines the feasible search space of the original left depth
signal. Similarly, a sequence $\mathcal{S}^r(\vec{b})$'s for blocks in the 
right map can also be deduced. 
Collectively, $\mathcal{S}^l(\vec{b})$'s and $\mathcal{S}^r(\vec{b})$'s 
are interpreted as two redundant descriptions of the same 3D scene. We 
next describe a POCS-like iterative procedure to enhance the precision of
reconstructed depth maps by considering both descriptions jointly.

\section{POCS-Inspired Iteration}
\label{sec:algo}

We perform the projection from left view to right view starting with an initial 
reconstructed left depth signal. The projection (composed of the following two 
steps) is alternated between the two views until convergence.

\subsection{View-to-view Projection}

In this step, we first back-project pixels of a
reconstructed left depth map to the 3D space. 
In particular, for a camera with intrinsic matrix $\vec{K}_{3\times 3}$ and
extrinsic matrix $\vec{E}_{3\times 4}$, a pixel with depth $d$ at image
coordinate $(r,c)$ corresponds to a 3D voxel at world coordinate $(x,y,d)$
that satisfies: $(c, r, 1)^\intercal = \alpha\cdot\vec{K}\cdot\vec{E}\cdot
(x,y,d,1)^\intercal$ for some scalar $\alpha$. The obtained 3D voxel is then
re-projected to the right view based on the right camera parameters. Note that
in general, the projected points to the right view do not land on 2D grid
points of the right depth map. Thus, to interpolate an updated right depth at
$(r,c)$, we perform a simple edge-adaptive linear interpolation using two
pixels $p_1$ and $p_2$ projected from the left view (assuming two views are
rectified). $p_1$ and $p_2$ are respectively the projected pixels whose depth
values are: 1) close to initial right depth value at $(r,c)$; 2) minimal among
projected depth values in range $(r, (c-1, c))$ and $(r, (c, c+1))$
respectively.
After interpolation, we further apply a bilateral filter~\cite{tomasi98} to
remove noisy pixels.

\subsection{Coefficient Clipping}

This step is a standard projection on convex sets: we transform each depth pixel block
in the right view to DCT domain, and clip a transform coefficient 
to its nearest boundary value if it is out of range. It is to 
ensure that the quantization constraints in (\ref{eq:quan}) are satisfied.

After convergence, the original transform coefficient vectors
$\{\vec{y}(\vec{b})\}$ becomes modified ones $\{\vec{y^*}(\vec{b})\}$. The
output depth maps are then obtained using inverse transform.


\section{Preliminary Results}
\label{sec:results}
The test sequences are obtained from the New Tsukuba Stereo Dataset\footnote{http://www.cvlab.cs.tsukuba.ac.jp/dataset/tsukubastereo.php}. Standard decoded depth maps are denoted as $\vec{I}^{std}$, while the outputs of proposed method are $\vec{I}^{our}$. Corresponding quality metrics are calculated by:
 
\begin{equation}
Q_{std} = g(\vec{I}_l^{std}, \vec{I}_r^{std}),\ \ \ Q_{our} = g(\vec{I}_l^{our}, \vec{I}_r^{our})
\end{equation}
where function $g(\vec{I}_1,\vec{I}_2) = (\text{PSNR}(\vec{I}_1, \vec{I}_l) + \text{PSNR}(\vec{I}_2, \vec{I}_r))/2$ calculates the average PSNR with regard to uncompressed 8-bit depth maps $\vec{I}_l$ and $\vec{I}_r$.

Note that bilateral filter is used in our proposed method. As a comparison, we
also apply bilateral filter directly on $\vec{I}^{std}$, resulting in smoothed
depth map $\vec{I}^{smo}$, whose average PSNR, $Q_{smo}$, is calculated
similarly.

\begin{figure}[ht]
\centering
\subfigure[left view]{\includegraphics[width = 0.24\textwidth]{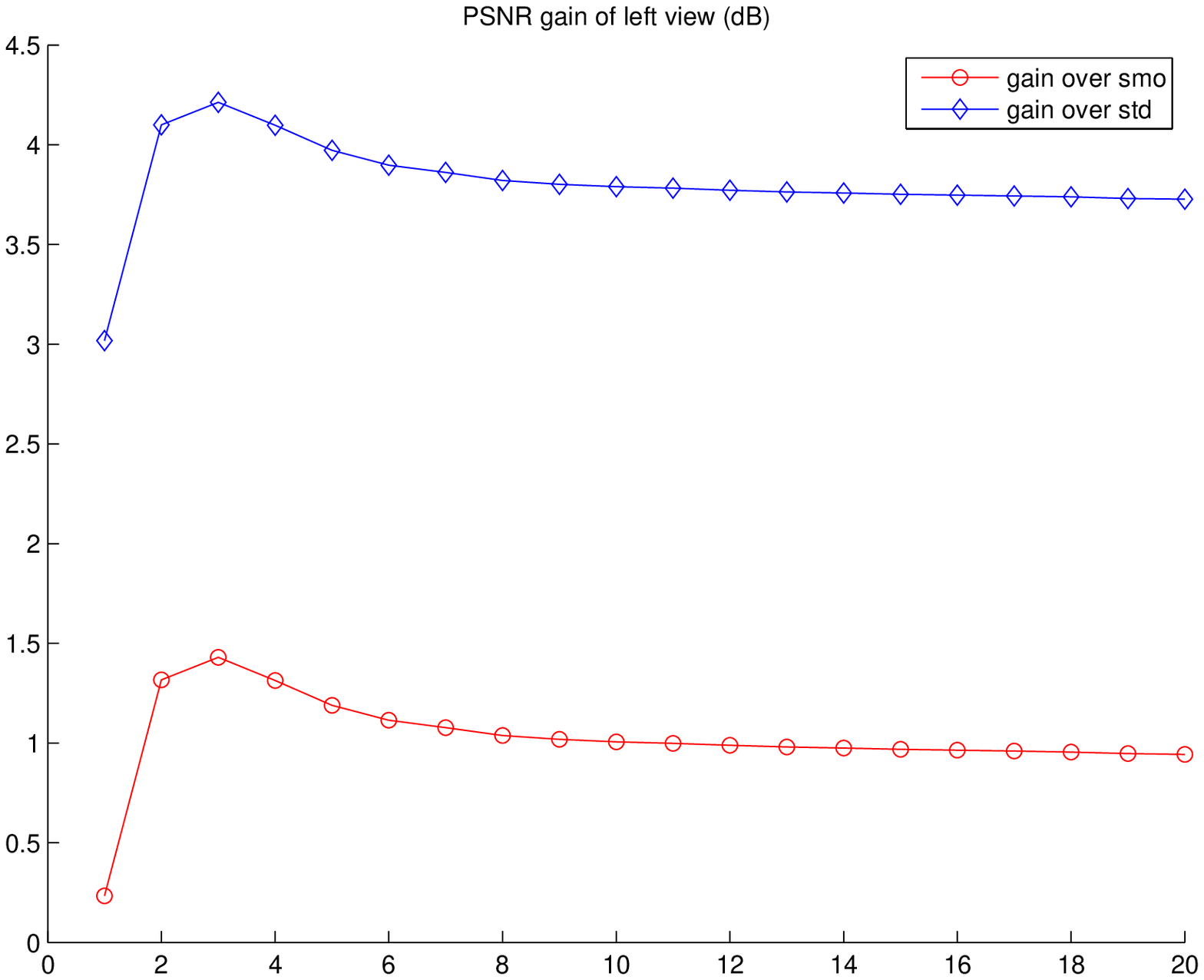}}
\subfigure[right view]{\includegraphics[width = 0.24\textwidth]{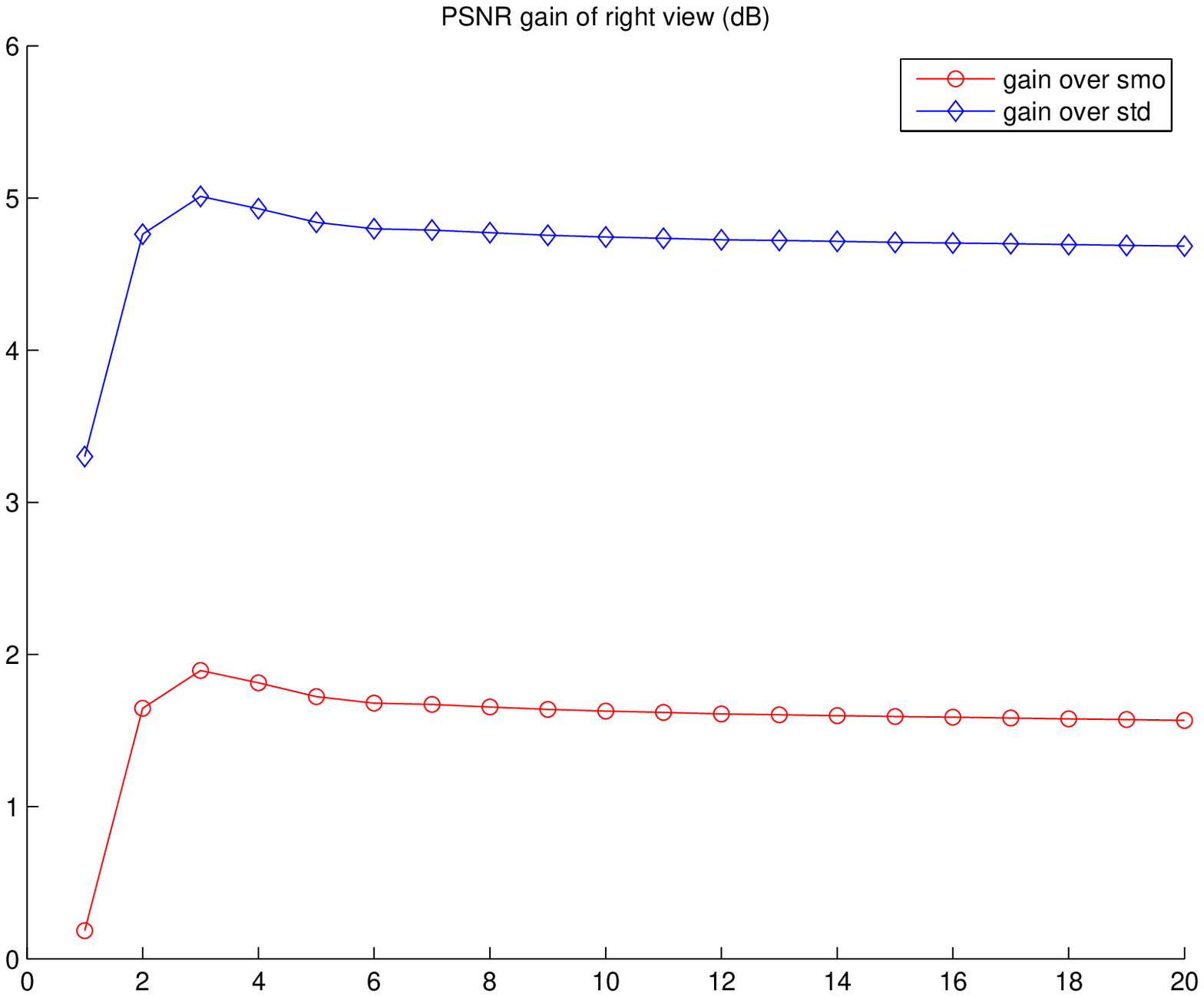}}
\caption{PSNR gain in dB. X-axis is the iteration index.}
\label{fig:psnr}
\end{figure}

\begin{figure}[ht]
\centering
\subfigure[Ground-truth $\vec{I}_l$]{\includegraphics[width = 0.24\textwidth]{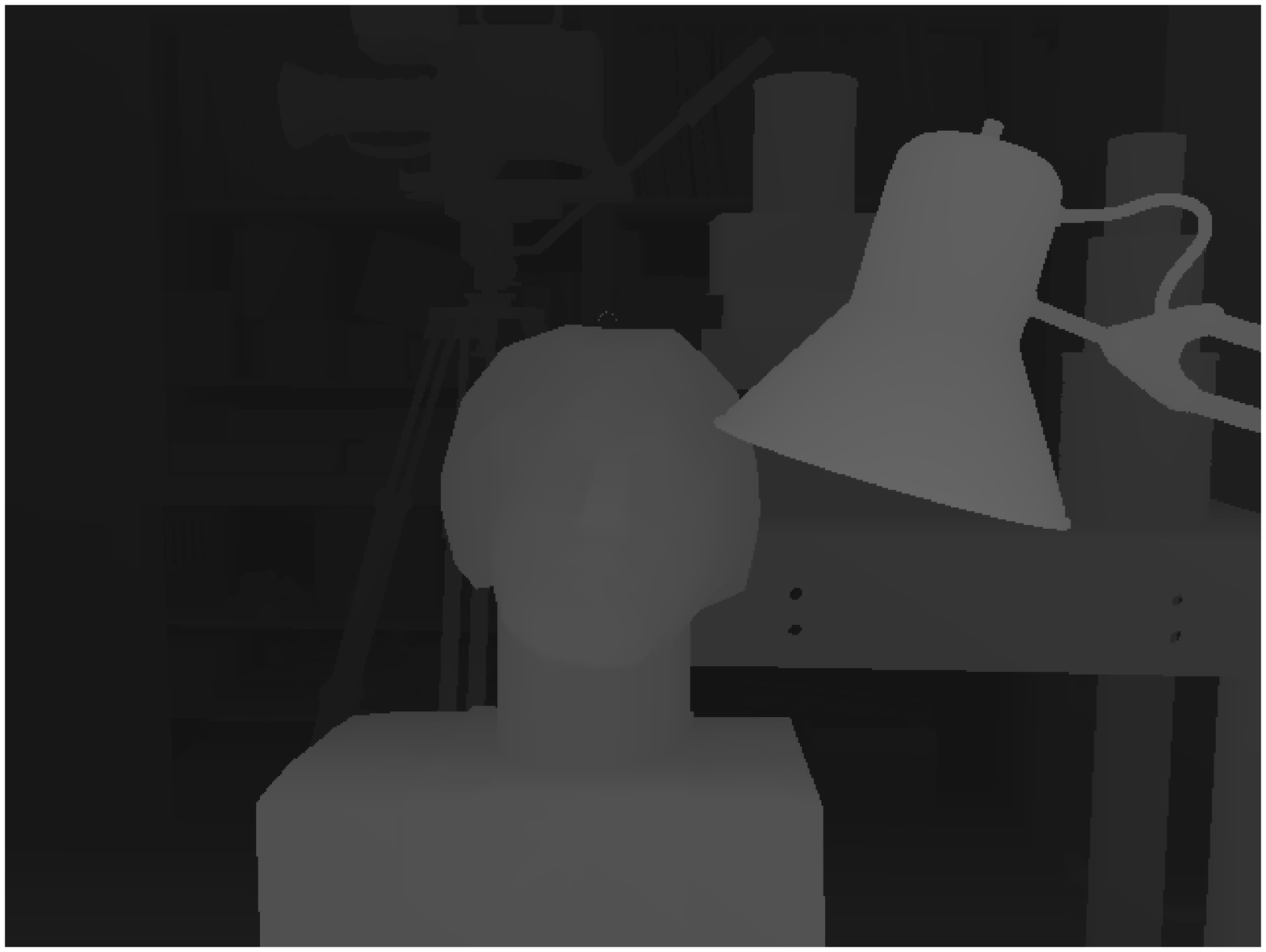}}
\subfigure[$|\vec{I}_l^{our}-\vec{I}_l|$]{\includegraphics[width = 0.24\textwidth]{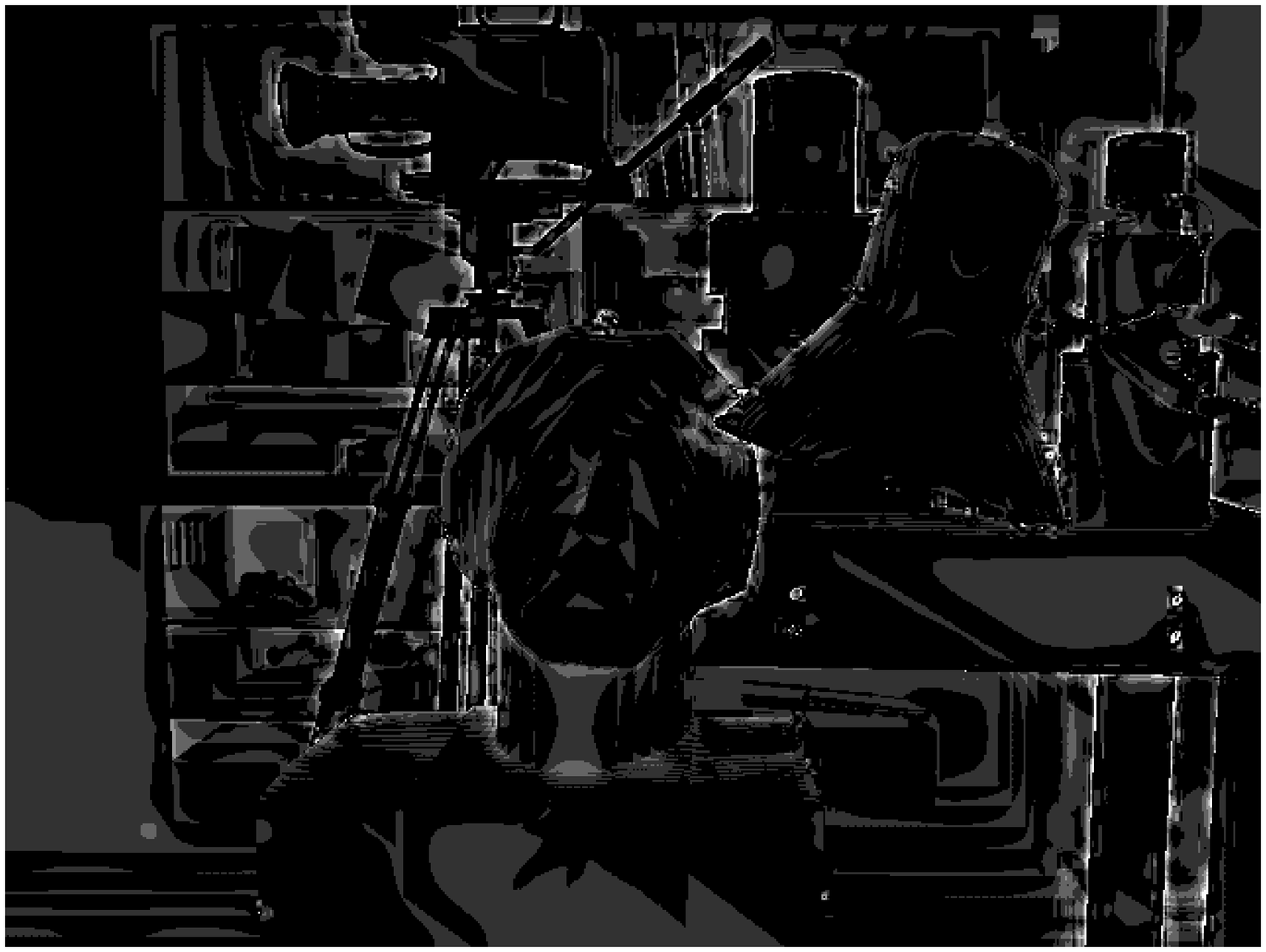}}
\subfigure[$|\vec{I}_l^{smo}-\vec{I}_l|$]{\includegraphics[width = 0.24\textwidth]{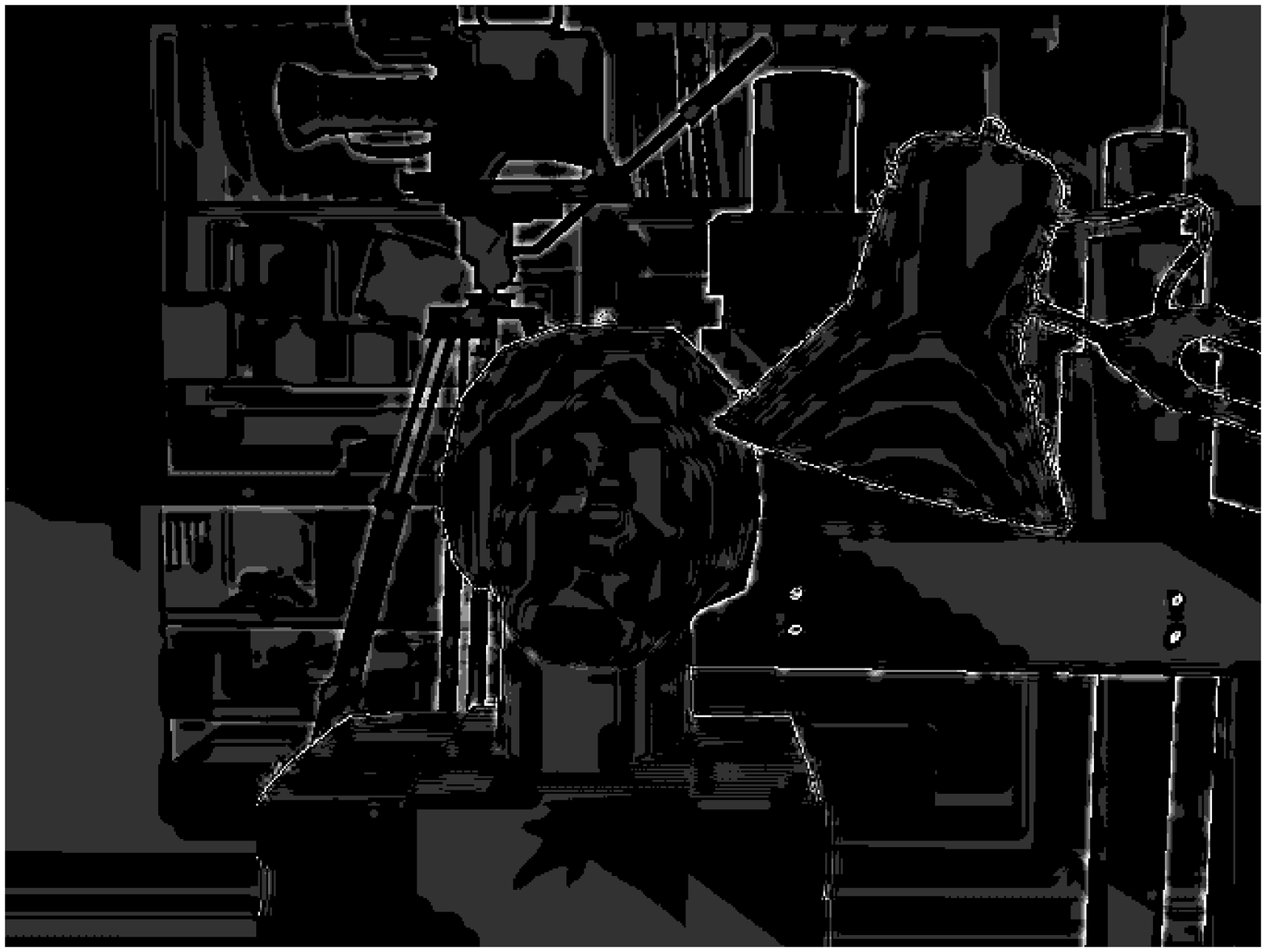}}
\subfigure[$|\vec{I}_l^{std}-\vec{I}_l|$]{\includegraphics[width = 0.24\textwidth]{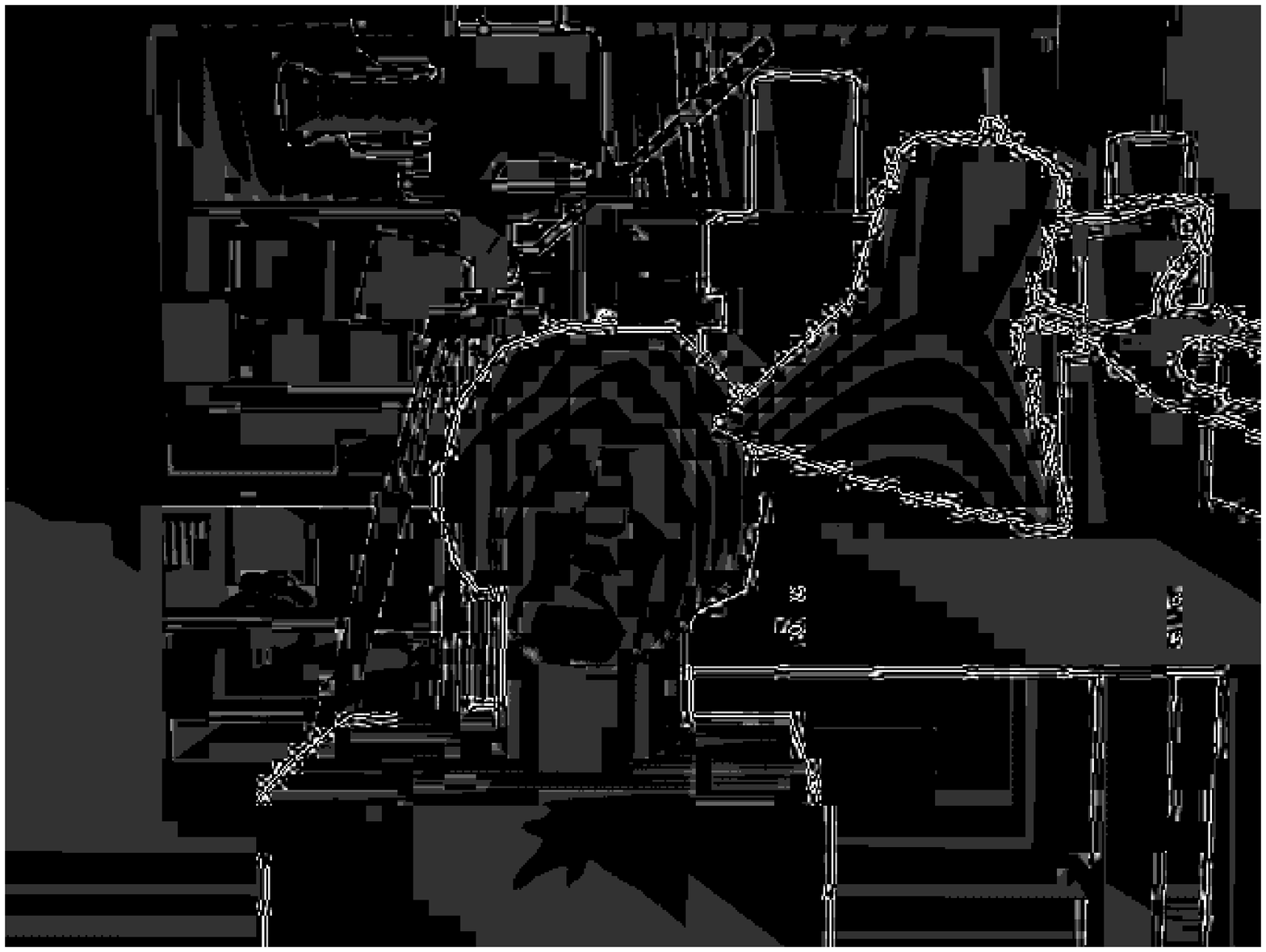}}
\caption{Sample error maps of the left view.}
\label{fig:subjective}
\end{figure}

Fig.~\ref{fig:psnr} shows the PSNR gain of proposed method over comparing ones: $Q_{our}-Q_{smo}$ and $Q_{our}-Q_{std}$. Error maps are also shown in Fig.~\ref{fig:subjective}. Overall, proposed method achieves encouraging performance: $Q_{our}>Q_{smo}>Q_{std}$.


Note that convergence of our POCS-like method is empirical, and that the convergence point is typically not the global optimal one, see Fig.~\ref{fig:psnr}. Proof of convergence and study of the optimal number of iterations are left for future work.

\section{Conclusion}
\label{sec:conclude}
In this paper, we introduce a POCS-like method to enhance the bit-precision for decoded multi-view depth maps. The experimental results show that proposed method significantly outperforms comparing methods. 


\bibliographystyle{IEEEtran}
\bibliography{ref}

\begin{thebibliography}{1}
\providecommand{\url}[1]{#1}
\csname url@samestyle\endcsname
\providecommand{\newblock}{\relax}
\providecommand{\bibinfo}[2]{#2}
\providecommand{\BIBentrySTDinterwordspacing}{\spaceskip=0pt\relax}
\providecommand{\BIBentryALTinterwordstretchfactor}{4}
\providecommand{\BIBentryALTinterwordspacing}{\spaceskip=\fontdimen2\font plus
\BIBentryALTinterwordstretchfactor\fontdimen3\font minus
  \fontdimen4\font\relax}
\providecommand{\BIBforeignlanguage}[2]{{%
\expandafter\ifx\csname l@#1\endcsname\relax
\typeout{** WARNING: IEEEtran.bst: No hyphenation pattern has been}%
\typeout{** loaded for the language `#1'. Using the pattern for}%
\typeout{** the default language instead.}%
\else
\language=\csname l@#1\endcsname
\fi
#2}}
\providecommand{\BIBdecl}{\relax}
\BIBdecl

\bibitem{wenxiu13}
W.~Sun, G.~Cheung, P.~A. Chou, D.~Florencio, C.~Zhang, and O.~C. Au,
  ``Rate-distortion optimized {3D} reconstruction from noise-corrupted
  multiview depth videos,'' in \emph{IEEE International Conference on
  Multimedia \& Expo}, San Jose, USA, 2013.

\bibitem{ivmsp13}
P.~Wan, G.~Cheung, P.~A. Chou, D.~Forencio, C.~Zhang, and O.~C. Au, ``Precision
  enhancement of {3D} surfaces from multiple quantized depth maps,'' in
  \emph{11th IEEE IVMSP Workshop: 3D Image/Video Technologies and
  Applications}, Seoul, Korea, 2013.

\bibitem{phil99_pocs}
P.~Chou, S.~Mehrotra, and A.~Wang, ``Multiple description decoding of
  overcomplete expansions using projections onto convex sets,'' in \emph{In
  Proceedings of Data Compression Conference}, 1999, pp. 72--81.

\bibitem{tomasi98}
C.~Tomasi and R.~Manduchi, ``Bilateral filtering for gray and color images,''
  in \emph{Proceedings of the IEEE International Conference on Computer
  Vision}, Bombay, India, 1998.

\end{thebibliography}

\end{document}